\documentclass{article}
\usepackage{spconf,amsmath,graphicx,stfloats,cite}

\title{END-TO-END CODE-SWITCHING ASR FOR LOW-RESOURCED LANGUAGE PAIRS}
\name{Xianghu Yue \textsuperscript{1,2}, Grandee Lee \textsuperscript{2}, 
      Emre Y{\i}lmaz \textsuperscript{2}, Fang Deng \textsuperscript{1}, Haizhou Li \textsuperscript{2}}

\address{\textsuperscript{1}Beijing Institute of Technology, Beijing, China \\
	     \textsuperscript{2}National University of Singapore, Singapore \\
	     \{xianghu.yue, grandee.lee\}@u.nus.edu, fangdeng@bit.edu.cn, \{emre, haizhou.li\}@nus.edu.sg}

\begin{document}
%
\maketitle
\begin{abstract}
Despite the significant progress in end-to-end (E2E) automatic speech recognition (ASR), E2E ASR for low resourced code-switching (CS) speech has not been well studied. In this work, we describe an E2E ASR pipeline for the recognition of CS speech in which a low-resourced language is mixed with a high resourced language. Low-resourcedness in acoustic data hinders the performance of E2E ASR systems more severely than the conventional ASR systems.~To mitigate this problem in the transcription of archives with code-switching Frisian-Dutch speech, we integrate a designated decoding scheme and perform rescoring with neural network-based language models to enable better utilization of the available textual resources. We first incorporate a multi-graph decoding approach which creates parallel search spaces for each monolingual and mixed recognition tasks to maximize the utilization of the textual resources from each language. Further, language model rescoring is performed using a recurrent neural network pre-trained with cross-lingual embedding and further adapted with the limited amount of in-domain CS text. The ASR experiments demonstrate the effectiveness of the described techniques in improving the recognition performance of an E2E CS ASR system in a low-resourced scenario.

\end{abstract}
\section{Introduction}
\label{sec:intro}
As multilingualism is becoming more common in today's globalized world \cite{Cloin}, there has been increasing interest in code-switching (CS) automatic speech recognition (ASR)~\cite{sitaram2019survey}. Code-switching refers to the phenomenon where two languages are spoken in contact within one utterance \cite{Peter}. Code-switching, such as Mandarin-English \cite{Dau}, Spanish-English \cite{Alfredo} and Hindi-English \cite{Anik}, is commonly practiced in multi-lingual societies.

Traditionally, an ASR system consists of several components including acoustic model, pronunciation and language model that are separately trained and optimized with different objectives, thus building an ASR system needs specialized expertise in the field. Various end-to-end (E2E) ASR approaches are emerging quickly because of its simplicity compared to the traditional ASR architecture. An E2E system predicts phones or characters directly from acoustic information without predefined alignment. Some notable architectures include connectionist temporal classification (CTC) \cite{Graves2012}, attention based encoder-decoder networks \cite{Jan,Cho}, and recurrent neural network (RNN) transducers \cite{Graves2014}. More recently, hybrid E2E systems have been successfully implemented and applied to common ASR benchmarks~\cite{watanabe2017}. These E2E models have been successfully used in monolingual and multilingual ASR systems by achieving promising results on various benchmarks \cite{Battenberg,Chan2016,Chan2017,Jinyu, shan}. 

E2E ASR approaches enable lexicon-free recognition which is a key advantage over traditional hybrid hidden Markov model/deep neural networks (HMM/DNN) approaches in low-resourced settings, since there are many low-resourced languages without an available pronunciation lexicon. However, there is very limited work done for recognizing CS speech using E2E techniques, especially for low-resourced language pairs. This is mainly due to the fact that low-resourcedness in acoustic data hinders the performance of E2E CS ASR more severely than the conventional ASR system. Hiroshi \cite{Hiroshi} built an encoder-decoder based E2E ASR system that can recognize the mixed-language speech. However, the work relies on training data that is generated from monolingual datasets, rather than natural code-switching speech. Kim \cite{Kim} and Toshniwa \cite{Toshniwa} both used encoder-decoder model to build multilingual E2E ASR, but their systems cannot deal with CS scenario. Li \cite{Li} incorporate a frame-level language identification (LID) model to linearly adjust the posteriors of an E2E CTC model for the high-resourced Mandarin-English language pair.

In this paper, we integrate a designated decoding scheme and a code-switch language model (LM) rescoring scheme to mitigate this problem in our recognition scenario, namely transcripts of archives with CS Frisian-Dutch speech in which Frisian is a low-resourced language and Dutch is a high-resourced language. The code-switch LM \cite{lee2019} is a recurrent neural network (RNN) that is trained with cross-lingual embedding and adapted to maximize the use of the available textual resources.~The decoding scheme provides a new multi-graph back-end for E2E CS ASR in which parallel search spaces are employed for monolingual and mixed recognition subtasks. The code-switch RNN LM can both preserve the cross-lingual correspondence derived from larger monolingual textual resources and leverage the low-resourced language on the high-resourced language at the same time.

The rest of this paper is organized as follows. Section 2 introduces the E2E CTC acoustic model. The incorporated multi-graph decoding strategy and CS RNN LM rescoring are described in section 3 and 4 respectively. We describe the experimental setup in section 5 and then present and discuss the results provided by the described E2E ASR pipeline in section 6.

\section{End-to-End CTC Acoustic Model}
\label{sec:acoustic}
Unlike in the traditional hybrid HMM-DNN system, an E2E CTC acoustic model is not trained using frame-level labels with respect to the cross-entropy (CE) criterion. Instead, a CTC model learns the alignments automatically between speech frames and their label sequences, i.e., phone sequences, by adopting the CTC objective. It predicts the conditional probability of the label sequence by summing over the joint probabilities of the corresponding set of CTC symbol sequences. The CTC framework has the output independent assumption that CTC symbols are conditionally independent at each frame, which may be more desirable for dealing with CS speech (though less accurate in general) as the current output does not explicitly depend on previous outputs~\cite{Li}. 
The conditional probability of the whole label sequence is:
\begin{equation}
P(\mathbf{z}|\mathbf{x}) = \sum\limits_{\mathbf{\pi}\in {{\mathcal{B}}^{ - 1}}(\mathbf{z})}{P(\mathbf{\pi}|\mathbf{x})} = \sum\limits_{\pi :\pi \in {Z^{'}} ,{\mathcal{B}}({\pi _{1:T}}) = \mathbf{z}} {\prod\limits_{t = 1}^T {y_{{\pi _t}}^t} }
\label{euq1}
\end{equation}
where $\mathbf{z} = ({z_1}, \ldots, {z_u}, \dots, {z_U})$ denotes a phone label sequence containing $U$ phones, $z \in Z$ and $Z$ is the phone set. $\mathbf{x} = ({x_1}, \ldots, {x_t}, \dots, {x_T})$ denotes a sequence of $T$ speech frames, with t being the frame index. The length of $z$ is constrained to be no greater than the length of the utterance, i.e., $U \leq T$. $\pi _{1:T} = ({\pi_1}, \ldots ,{\pi_t}, \dots, {\pi_T})$ is an output symbol sequence at frame level, named CTC \textsl{path}. Each output symbol $\pi \in Z^{'}$ and $Z^{'} = Z \cup {blank}$. $blank$ is a special label in the CTC framework, which maps frames and labels to the same length. $\mathcal{B}$ is a multiple-to-one mapping with first removing the repeated labels and then all $blank$ symbols from the paths. ${y_{{\pi _t}}^t}$ is the posterior probability of output symbol ${\pi _t}$ at time $t$. Equation (\ref{euq1}) can be efficiently evaluated and differentiated using forward-backward algorithm \cite{Graves2006}. Given training utterances, the acoustic model networks are trained to minimize the CTC objective function:
\begin{equation}
\mathcal{L} = - \sum\limits_{k = 1}^Q {\ln (P(\mathbf{z_k}|\mathbf{x_k}))}
\end{equation}
where $k$ is the index of training utterances and $Q$ is the total number.

\section{Multi-graph Decoding Strategy}
\label{sec:decoding}
In modern ASR architectures, weighted finite-state transducers (WFST) are used to integrate different knowledge sources and perform search space optimization to achieve the best search efficiency using highly-optimized FST libraries such as OpenFST \cite{Mohri,Allauzen}. In E2E CTC ASR framework \cite{Miao}, individual components, containing CTC labels, lexicons, and N-gram language models, are encoded into three individual WFSTs and then composed into a comprehensive search graph that encodes the mapping from a CTC symbol sequence emitted from the speech frames to a sequence of words. The search space is represented as $ \textbf{T} \circ \textbf{L} \circ \textbf{G} $ in the Eesen toolkit \cite{Miao}, where $\textbf{T}$ is a \textit{token} WFST that maps a sequence of frame-level symbols to a single lexicon unit, $\textbf{L}$ is a lexicon WFST that encodes the mapping from sequences of lexicon units to words, and $\textbf{G}$ is a grammar WFST that encodes the word sequences information in N-gram language model. Thus, using WFST-based decoding framework, we can incorporate different word-level language model efficiently to make full use of the available textual resources and overcome the imbalance in acoustic data between low-resourced and high-resourced language in our CS scenario.

In our previous work \cite{Emre2019}, Yilmaz et al. proposed a multi-graph decoding strategy which creates parallel search spaces for each monolingual and bilingual recognition tasks for the conventional CS ASR system. This strategy can be easily extended to E2E CTC ASR system to address the above-mentioned data imbalance problem. For the multi-graph decoding strategy, we use the union operation to create a larger graph with parallel bilingual and monolingual (Frisian and Dutch) subgraphs. The parallel graphs used during decoding are characterized by the incorporated language model component, as they share the same token ($\textbf{T}$) and lexicon ($\textbf{L}$) components. This approach has been shown to outperform standard LM interpolation~\cite{Emre2019}, that makes effective use of the text resources of the high-resourced language by creating three different search spaces with an identical acoustic model (AM). Monolingual and code-mixed utterances are decoded using best-matching subgraph, yielding improved monolingual recognition performance on the high-resourced language without any accuracy loss on the code-mixed utterances.

\section{CS RNN Language Modeling}
\label{sec:language}
In language modeling, we face data sparsity both in terms of availability of CS corpus, and scarcity of CS occurrences in the corpus. To address these problems, we propose a two-step approach to language modeling. Firstly, in terms of data augmentation, we boost the size of CS corpus by synthetically generating CS text using a well-trained long short-term memory (LSTM) language model. Similar techniques are also proposed in \cite{grandee2019,Emre_data}. However, in \cite{grandee2019} a sentence level aligned parallel corpus is available, thus synthetic CS data can be generated based on word or phrase alignment between the parallel sentences and guided by linguistic rules. Unlike \cite{grandee2019}, we lack a parallel corpus, thus we cannot explicitly establish the word-level cross-lingual correspondence between the two languages. This motivates the second step of our language model, i.e., to find the cross-lingual mapping of the monolingual word embeddings using an unsupervised self-learning method proposed by \cite{artetxe-etal-2018-robust}. The method finds the mapping functions $W_M, W_N$ that maximize the cosine similarity between the monolingual embeddings of source language $M$ and target language $N$, based on an iteratively learned dictionary $D$:
\begin{equation}
\arg\max_{W_M, W_N}\sum_{i,j \in D} (M_{i}W_M)\cdot(N_jW_N)
\end{equation} 
$i, j$ are paired entries in the dictionary that represent a translation pair and $M_i, N_j$ are the respective monolingual embeddings. Since the transformation matrices and embeddings are length normalized, cosine similarity is optimized.
Thus, the method explicitly aligns the word based on the monolingual distributional property and projects both monolingual embedding into the same embedding space. Resultant word embeddings of the related words in both languages are grouped together and at the same time, monolingual syntactic information is preserved \cite{lee2019}. 

\begin{equation}
y_{k} = LSTM(w_k)
\vspace{-2mm}
\end{equation} 
\begin{equation}
p_k = \frac{e^{y_k}}{\sum_{j=1}^{V}e^{y_j}}
\label{eqn_5}
\vspace{-2mm}
\end{equation}
\begin{equation}
Loss = -\frac{1}{Q-1}\sum_{k=1}^{Q-1}Y_{k+1}\ln(p_k)
\label{eq6}
\end{equation}
This pre-trained cross-lingual embedding is used to initialize our neural language model and the embedding layer is fixed during training. The output $y_k$ from LSTM with the current word embedding $w_k$ is passed through a softmax function Eq. (\ref{eqn_5}) to form a distribution $p_k$ over the total vocabulary V,  which represents the next word probability. The loss function is the cross-entropy between the true target $Y_{k+1}$ and $p_k$ in Eq. (\ref{eq6}), where Q is the number of words in the corpus. By freezing the embedding layer, we aim to preserve the cross-lingual correspondence derived from larger monolingual corpora and let the low-resourced language leverage on the resource rich language.

\section{Experimental Setup}
\label{sec:majhead}
\subsection{Datasets}
\label{ssec:corpora}
The experiments are conducted on the low-resourced Frisian-Dutch CS corpus from the FAME! project, this project aims to develop a spoken document retrieval system for the disclosure of the archives of Omrop Frysl\^{a}n (Frisian Broadcast) covering a large time span and a wide variety of topics which contain monolingual Dutch and Frisian speech as well as code-mixed Frisian-Dutch speech. Further details can be found in \cite{Emre_cs}. It is worth mentioning that proposed approaches can also be applied to other low-resourced language pairs and scenarios with more than two languages as in~\cite{yilmaz2018_3}.  

The training data used in the experiments are summarized in Table \ref{tab:acdata}. Both monolingual and CS data is used for acoustic model training, since monolingual acoustic data augmentation has been shown to improve the CS ASR on both monolingual and code-mixed test utterances~\cite{Emre2018}. The manually annotated CS data is from the FAME corpus containing 8.5 hours and 3 hours of orthographically transcribed speech from Frisian (fy) and Dutch (nl) speakers respectively. The ‘Frisian Broadcast’ data containing 125.5 hours of automatically transcribed speech data extracted from the target broadcast archive. Monolingual Dutch data comprises 442.5 hours Dutch component of the Spoken Dutch Corpus (CGN) \cite{Oostdijk} that contains diverse speech materials including conversations, interviews, lectures, debates, read speech and broadcast news. The development and test sets consist of 1 hour of speech from Frisian speakers and 20 minutes of speech from Dutch speakers each. The sampling frequency of all speech data is 16 kHz.
\begin{table}[!t]
\centering
\caption{Acoustic data composition used for CTC AM training (in hours)}
\vspace{0.25cm}
\begin{tabular}{| l | c | c | c | c | c |}
\hline
Traning data               &  Annot.        & Frisian       & Dutch       & Total \\
\hline 
(1) FAME                  &  Manual         & 8.5           & 3.0         &  11.5 \\
\hline
(2) Frisian Broad.         &   Auto.        & \multicolumn{2}{c|}{125.5}  &  125.5 \\
\hline
(3) CGN-NL                 &  Manual        &   -          & 442.5      &  442.5 \\
\hline
\end{tabular}
\label{tab:acdata}
\end{table}

\subsubsection{Text data}
Bilingual text corpus (107M words) consisting of generated CS text (61M words), monolingual Frisian text (37M words) and monolingual Dutch text (9M words) are used for training the baseline CS LM. The transcripts of the FAME training data is the only source of CS text containing 140k words and textual data augmentation techniques described in \cite{Emre2018} have been applied to increase the amount of CS text. The Frisian text is extracted from monolingual resources such as Frisian novels, news and Wikipedia articles. The Dutch text is extracted from the transcripts of the CGN speech corpus. We use the larger monolingual subset (300M words) of the NLCOW text corpus\footnote{http://corporafromtheweb.org} together with Dutch text (9M words) which is used in baseline CS LM to train larger Dutch LM and create larger monolingual Dutch graph.

\subsection{Implementation details}
\label{ssec:implement}
All the recognition experiments are performed in the Eesen E2E CTC ASR toolkit~\cite{Miao}. The 3-fold data augmentation~\cite{ko2015} is applied to the in-domain acoustic training data, i.e., (1) and (2) in Table~\ref{tab:acdata}. The acoustic model is a 6-layer bidirectional LSTM with 640 hidden units trained without predefined alignment. The 40-dimensional filterbank features with their first and second-order derivatives are stacked using 3 contiguous frames to form 360-dimensional spliced features as inputs. The features are normalized via mean subtraction and variance normalization on a per-speaker basis. The learning rates starts at 0.00004 and remains unchanged until the drop of label error rate on validation set between two consecutive epochs falls below 0.5\%. From then on, the learning rate is halved at the subsequent epochs. The conventional ASR system is trained using the Kaldi ASR toolkit \cite{kaldi}. A context-dependent Gaussian mixture model-hidden Markov model (GMM-HMM) system is firstly trained using MFCC including the deltas and deltas-deltas to obtain the alignments. Then these alignments are used for training a TDNN-LSTM acoustic model (1 standard, 6 time-delay and 3 LSTM layers) with LF-MMI \cite{lf-mmi} criterion using 40-dimensional MFCC as features combined with i-vectors for speaker adaptation. 

The language models used in the first pass ASR decoding are standard bilingual 3-grams with interpolated Kneser-Ney smoothing. The baseline RNN LM with gated recurrent units (GRU) has 400 hidden units and is trained using noise contrastive estimation\footnote{https://github.com/yandex/faster-rnnlm} for lattice rescoring. The CS RNN LM with the same architecture is adapted to the CS transcripts to reduce the mismatch. The adaptation is performed at the last 5 epochs while following the overall learning rate decay of 0.8. In summary, we have 7 LMs: (1) baseline CS LM (cs) trained on the bilingual text (107M), (2) baseline monolingual Frisian LM (fy) trained on monolingual Frisian text (37M), (3) baseline monolingual Dutch LM (nl) trained on monolingual Dutch text (9M), (4) larger monolingual Dutch LM (nl++) trained on 309M words, (5) interpolated LM (interp-nl++) with the interpolation between cs LM and nl++ LM, whose interpolation weight yields the lowest perplexity on the development set, (6) baseline RNN LM trained on the corresponding bilingual text (107M) using 1 layer LSTM with 400 hidden units, (7) CS RNN LM trained using the similar parameters. The RNN LM weight for rescoring is 0.75. The first five LMs are used in the conventional signal-graph E2E ASR systems for comparison with the corresponding multi-graph decoding systems using the same amount monolingual and bilingual text. The perplexities of the baseline CS and the Dutch LMs on the monolingual Dutch component of the development and test set are shown in Table \ref{tab:perp}, the perplexities of two RNN LMs on development and test set show that CS RNN LM has a lower perplexity than its baseline in Table~\ref{tab:perp1}. 

\begin{table}[!t]
\centering
\caption{Perplexities obtained on the Dutch component of the development and test set using different LMs}
\addtolength{\tabcolsep}{-2pt}
\vspace{0.25cm}
\begin{tabular}{| l | c | c | c | c |}
\hline
LM        	        & Total \#  words  & Dev.      & Test \\
\hline \hline
Baseline CS LM      & 107M                  & 188    &  197 \\
\hline
interp-nl++ LM       & 416M                & 176    &  182 \\
\hline \hline
Baseline NL LM       & 9M                  & 150    &  151 \\
\hline
nl++ LM              & 309M                & 123    &  119 \\
\hline

\end{tabular}
\label{tab:perp}
\vspace{-0.5cm}
\end{table}
 
\begin{table}[!t]
\centering
\caption{Perplexities obtained on the different components of development and test transcripts using different LMs}
\vspace{0.25cm}
\begin{tabular}{| l | c c c  | c c c |}
\hline
 {} & \multicolumn{3}{c|}{Dev.} & \multicolumn{3}{c|}{Test} \\
\hline
{}  & fy & nl & cs & fy & nl & cs  \\
\hline
 {3-gram LM} & 158 & 191 & 272 & 138 & 189 & 227\\
 \hline
 {Base. RNN LM} & 205 & 187 & 330 & 177 & 177 & 283\\
\hline
 {CS RNN LM} & 183 & 164 & 296 & 159 & 156 & 257\\
\hline 
\end{tabular}
\label{tab:perp1}
\vspace{-0.3cm}
\end{table}

\begin{table}[!t]
\centering
\caption{WER (\%) obtained on the monolingual utterances in the development and test set of the FAME Corpus}
\addtolength{\tabcolsep}{-1.7pt}
\vspace{0.2cm}
\begin{tabular}{| l | c | c c | c c |}
\hline
\multicolumn{2}{|c|}{} & \multicolumn{2}{c|}{Dev.} & \multicolumn{2}{c|}{Test} \\
\hline
\multicolumn{2}{|c|}{} & fy & nl & fy & nl \\
\hline
\multicolumn{2}{|c|}{\# of Frisian words} & 9190 & 0 & 10753 & 0 \\
\hline
\multicolumn{2}{|c|}{\# of Dutch words} & 0 & 4569 & 0 & 3475 \\
\hline\hline
ASR System  & Graph  & \multicolumn{2}{c|}{} & \multicolumn{2}{c|}{} \\
\hline \hline
Baseline CS ASR & cs  & 32.9      & 33.7       & 30.6    & 29.0    \\
\hline 
fy              & fy  & 32.5      & -          & 30.8    & -       \\
\hline
nl              & nl  &  -         & 33.6      & -       & 27.9     \\
\hline
nl++            & nl++  &  -       & \bf{30.1} & -       & \bf{25.9} \\
\hline
\end{tabular}
\label{tab:wer_ideal}
\vspace{-0.3cm}
\end{table}
\vspace{-0.2cm}
\subsection{ASR experiments}
\label{ssec:asr}
Four sets of ASR experiments are conducted to evaluate the performance of the proposed method. Firstly, the ASR performance of the baseline single-graph ASR systems using cs and interp-nl++ LMs are presented. Secondly, the results provided by the bi-graph systems using the cs graph together with one of the monolingual graphs, namely fy, nl and nl++, are presented. Thirdly, tri-graph decoding systems with varying monolingual graphs are evaluated. 

After finalizing the multi-graph decoding experiments, we present the RNN LM rescoring experiment performed to evaluate the performance of CS RNN LM on CS speech compared to a baseline RNN LM. For the rescoring of the multi-graph systems, graph identification tags are used to identify the graph used for the hypothesized ASR output and then the rescoring is performed with the corresponding RNN LM. The CS RNN LMs are trained on the same text data with the N-gram used in decoding. The monolingual Frisian and Dutch RNN LMs are trained on Frisian text corpora (fy, 37M) and the largest Dutch text corpora (nl++, 309M) respectively using the same parameters as the baseline and CS RNN LMs. The recognition results are reported separately for Frisian only (fy), Dutch only (nl) and code-mixed (fy-nl) utterances. The overall performance is also reported to use as an overall performance indicator. The recognition performance of the ASR system is quantified using the word error rate (WER).

\section{Results and Discussion}
\label{sec:results}
The recognition results obtained by using only monolingual graphs on the corresponding monolingual utterances are presented in Table \ref{tab:wer_ideal}. The ASR system using only Frisian (fy) graph gives similar recognition performance to the baseline CS system on monolingual Frisian utterances, which indicates that the latter CS system has the ability to recognize monolingual Frisian speech as well as a monolingual Frisian ASR system. For monolingual Dutch utterances, the performance by using only Dutch (nl) graph is slightly better than baseline CS system on the test set with a WER of 27.9\% compared to 29.0\%. Using the largest monolingual Dutch graphs nl++ yields a WER of 25.9\% on the Dutch utterances respectively, revealing that the performance of the baseline CS graph can be improved by using larger monolingual Dutch graph in a multi-graph decoding framework.

\begin{table*}[!t]
\centering
\caption{WER (\%) obtained on the development and test set of the FAME Corpus}
\addtolength{\tabcolsep}{-2pt}
\vspace{0.3cm}
\begin{tabular}{| l | c | c | c c c c | c c c c | c |}
\hline
\multicolumn{3}{|c|}{} & \multicolumn{4}{c|}{Dev.} & \multicolumn{4}{c|}{Test} & Total\\
\hline
\multicolumn{3}{|c|}{} & fy & nl & fy-nl & all & fy & nl & fy-nl & all & \\
\hline
\multicolumn{3}{|c|}{\# of Frisian words} & 9190 & 0 & 2381 & 11\,571 & 10\,753 & 0 & 1798 & 12\,551 & 24\,122\\
\hline
\multicolumn{3}{|c|}{\# of Dutch words} & 0 & 4569 & 533 & 5102 & 0 & 3475 & 306 & 3781 & 8883\\
\hline \hline
ASR System  & Graph(s) & Rescoring & \multicolumn{4}{c|}{} & \multicolumn{4}{c|}{} & \\
\hline \hline
Kaldi CS ASR & cs   & No    & 26.3 & 27.6 & 36.8 & 28.4 & 25.1 & 24.4 & 39.3 & 26.7 & 27.6 \\
\hline \hline
\multicolumn{3}{|c|}{\textit{Single-graph systems}} & \multicolumn{4}{|c|}{} & \multicolumn{4}{|c|}{} & \\
\hline \hline
Base. E2E CS ASR & cs   & No     & 32.9   & 33.7 & 42.6 & 34.9 & 30.6 & 29.0 & 42.4 & 31.8 & 33.4\\
\hline
Base. E2E CS ASR & cs   & Yes    & 31.6   & 32.8 & 42.1 & 33.9 & 29.6 & \bf{27.9} & \bf{40.7} & 30.7 & 32.3 \\
\hline
Base. E2E CS ASR & cs   & CS-RNN   & 30.4 & \bf{31.2} & \bf{41.0} & 32.5 & 29.0 & 28.6  & 41.2 & 30.6 & \bf{31.6} \\
\hline
interp-nl++  & cs-nl++ & No     & 32.6   & 32.3  & 42.3  & 34.3 & 30.7 & 28.7 & 42.6 & 31.8 & 33.1 \\
\hline
interp-nl++  & cs-nl++ & Yes    & 31.3   & 32.5  & 41.5  & 33.4 & 29.9 & 28.2 & 41.0 & 31.0 & 32.2  \\
\hline \hline
\multicolumn{3}{|c|}{\textit{Multi-graph systems}} & \multicolumn{4}{|c|}{} & \multicolumn{4}{|c|}{} & \\
\hline \hline
union-fy     & cs, fy  & No & 32.7 & 33.0 & 42.3 & 34.5 & 30.7 & 28.6 & 42.7 & 31.9 & 33.2 \\
\hline
union-nl     & cs, nl  & No & 32.7 & 32.5 & 42.8 & 34.4 & 30.6 & 28.0 & 42.6 & 31.6 & 33.0 \\
\hline
union-nl++   & cs, nl++& No & 32.8 & 30.1 & 42.5 & 33.8 & 30.6 & 26.7 & 42.5 & 31.4 & 32.6 \\
\hline
union-nl++   & cs, nl++& Yes & 31.9 & \bf{28.4} & \bf{42.1} & 32.8 & 29.7 & \bf{23.6} & \bf{42.4} & 30.1 & \bf{31.5} \\
\hline \hline
union-fy-nl  & cs, fy, nl  & No & 32.9 & 32.4 & 42.9 & 34.6 & 30.8 & 28.1 & 42.8 & 31.8 & 33.2 \\
\hline
union-fy-nl++& cs, fy, nl++& No & 32.9 & 30.1 & 42.8 & 33.9 & 30.8 & 25.6 & 43.1 & 31.3 & 32.5 \\
\hline
union-fy-nl++& cs, fy, nl++& Yes & 32.3 & 28.2 & 41.7 & 32.8 & 30.2 & 23.1 & 41.3 & 30.2 & 31.6 \\
\hline
union-fy-nl++& cs, fy, nl++& CS-RNN & 32.3 & \bf{28.2} & \bf{40.5} & 32.6 & 30.2 & \bf{23.1} & \bf{40.5} & 30.1 & \bf{31.4} \\
\hline
\end{tabular}
\label{tab:wer}
\end{table*}

The ASR results obtained using multi-graph decoding strategy and the CS RNN LM rescoring are presented in Table \ref{tab:wer}. The number of Frisian and Dutch words in each component of development and test sets are presented in the upper panel. Then two baseline results using single-graph systems (cs and interp-nl++) are shown in the middle panel. The results provided by an equivalent Kaldi~\cite{kaldi} ASR system with conventional architecture is also given as a reference. Compared to the baseline E2E CS ASR system, using the interpolated larger Dutch LM brings marginal improvements from 33.7\% (29.0\%) to 32.3\% (28.7\%) on the development (test) set. This indicates that using interpolated larger LM in single graph is ineffective in improving the accuracy on monolingual utterances. 

Finally, the ASR results provided by the multi-graph E2E ASR systems are presented in the bottom panel. According to these results, using an additional monolingual Frisian graph during the multi-graph decoding (union-fy and union-fy-nl) does not improve the ASR performance on the fy utterances, which is consistent with the previous results reported in~\cite{Emre2019}. Including the largest monolingual Dutch graph in the union-fy-nl++ system improves the ASR accuracy on nl utterances with a WER of 30.1\% (25.6\%), yielding a 10.7\% (11.7\%) relative WER reduction. 

For RNN LM rescoring, CS RNN LM provides absolute overall 0.7\% WER reduction from 32.3\% to 31.6\% over the baseline RNN LM in single-graph systems and 1.2\% (0.8\%) WER reduction on fy-nl utterances for the union-fy-nl++ system perhaps due to the fact that the CS RNN LM could preserve more cross-lingual information. The Dutch RNN LM (trained on 309M Dutch text corpora) provides the best WER of 28.2\% (23.1\%) on monolingual Dutch utterances, while the Frisian RNN LM (trained on 37M Frisian text) and the baseline RNN LM (trained on 107M bilingual text) give limited improvements on the corresponding subsets. Finally, the WER of E2E CTC ASR system is significantly reduced to 31.4\%.

\section{Conclusion}
In this paper, we propose an E2E CTC ASR pipeline for a CS scenario in which a low-resourced language is mixed with a high-resourced language. We first incorporate a multi-graph decoding strategy by creating parallel search spaces for monolingual and code-switching recognition tasks. Moreover, we perform language model rescoring using a recurrent neural network pre-trained with cross-lingual embedding and then adapted with the limited amount of in-domain code-switching text. For evaluating the effectiveness of the proposed pipeline, ASR experiments are conducted on the Frisian-Dutch CS speech, in which the target Frisian language is low-resourced with limited acoustic and textual resources while Dutch language is high-resourced. The experimental results demonstrate that the multi-graph decoding approach can improve monolingual Dutch recognition performance of an E2E CS ASR system without degradation in the CS performance. The adapted recurrent neural network language model further improves the performance on CS speech. Finally, the proposed pipeline gives 16.3\% (20.3\%) relative WER reduction on monolingual Dutch speech and absolute 2.1\% (1.9\%) WER reduction on code-switching speech.

\section{Acknowledgements}
This research is supported by the National Research Foundation Singapore under its AI Singapore Programme (Award Number: AISG-100E-2018-006). This research is also supported by the Agency for Science, Technology and Research (A*STAR) under its AME Programmatic Funding Scheme (Project \#A18A2b0046). This research is partially supported by the Key Program of National Natural Science Foundation of China (No.61933002) and the National Key Research and Development Program of China (No.2018YFB1309300).
\bibliographystyle{IEEEbib}
\bibliography{strings,refs_EY}

\end{document}